\documentclass[conference]{IEEEtran}
\IEEEoverridecommandlockouts
\usepackage{cite}
\usepackage[utf8]{inputenc}
\usepackage{comment}
\usepackage{array}
\usepackage{longtable}
\usepackage{xspace} 
\usepackage{color}
\usepackage{multirow}
\usepackage{amsmath}
\usepackage{mathtools}
\usepackage{graphicx}
\usepackage{tabularx}
\usepackage{verbatim}

\usepackage{enumitem,kantlipsum}

\usepackage{color, colortbl}
\newcolumntype{Y}{>{\centering\arraybackslash}X}
\newcolumntype{L}{>{\raggedleft\arraybackslash}X}

\definecolor{RRRR}{rgb}{0,0,0} 
\definecolor{RRRO}{rgb}{0.1,0.1,0.1} 
\definecolor{RROO}{rgb}{0.2,0.2,0.2}
\definecolor{ROOO}{rgb}{0.3,0.3,0.3}
\definecolor{OOOO}{rgb}{0.4,0.4,0.4}
\definecolor{OOOY}{rgb}{0.5,0.5,0.5}
\definecolor{OOYY}{rgb}{0.6,0.6,0.6}
\definecolor{OYYY}{rgb}{0.7,0.7,0.7}
\definecolor{YYYY}{rgb}{0.8,0.8,0.8}

\def\BibTeX{{\rm B\kern-.05em{\sc i\kern-.025em b}\kern-.08em
    T\kern-.1667em\lower.7ex\hbox{E}\kern-.125emX}}

\begin{document}
\title{Semantic-based Distance Approaches in Multi-objective Genetic Programming}

\author{
\IEEEauthorblockN{Edgar Galv\'an$^*$\thanks{$^*$Main and senior author.}}
\IEEEauthorblockA{{Naturally Inspired Computation Research Group} \\
{Department of Computer Science},
{Hamilton Institute}\\
Maynooth University, Ireland \\
edgar.galvan@mu.ie}
\and
\IEEEauthorblockN{Fergal Stapleton}
\IEEEauthorblockA{{Naturally Inspired Computation Research Group} \\
{Department of Computer Science},
{Hamilton Institute}\\
Maynooth University, Ireland \\
fergal.stapleton.2020@mumail.ie}

}


\maketitle

\begin{abstract}

Semantics in the context of Genetic Program (GP) can be understood as the behaviour of a program given a set of inputs and has been well documented in improving performance of GP for a range of diverse problems. There have been a wide variety of different methods which have incorporated semantics into single-objective GP. The study of semantics in Multi-objective (MO) GP, however, has been limited and this paper aims at tackling this issue.  More specifically, we conduct a comparison of three different forms of semantics in MOGP. One semantic-based method, (i) Semantic Similarity-based Crossover (SSC), is borrowed from single-objective GP, where the method has consistently being reported beneficial in evolutionary search.  We also study two other methods, dubbed (ii) Semantic-based Distance as an additional criteriOn (SDO) and (iii) Pivot Similarity SDO. We empirically and consistently show how by naturally handling semantic distance as an additional criterion to be optimised in MOGP leads to better performance when compared to canonical methods and SSC. Both semantic distance based approaches made use of a pivot, which is a reference point from the sparsest region of the search space and it was found that individuals which were both semantically similar and dissimilar to this pivot were beneficial in promoting diversity. Moreover, we also show how the semantics successfully promoted in single-objective optimisation does not necessary lead to a better performance when adopted in MOGP. 

\end{abstract}

\begin{IEEEkeywords}
Semantics, genetic programming, Multi-objective optimisation. 
\end{IEEEkeywords}

\section{Introduction}


Genetic Programming (GP)~\cite{koza_1994_genetic} is a form of Evolutionary Algorithm (EA) that uses genetic operations that are analogous to behavioural biology and evolve programs towards finding a solution to a problem. The range of problem domains for GP are wide and this form of EA has been found to be beneficial for problems with multiple local optima and for problems with a varying degree of complexity~\cite{eiben_2015_from}, making EAs ideal for highly complex problems including the automatic configuration of deep neural networks' architectures (an in-depth recent literature review in this emerging research area can be found in~\cite{galvan2020neuroevolution}). However despite the well documented effectiveness of canonical GP, there are well-known limitations of these methods, through the study of properties of encodings~\cite{DBLP:conf/cec/LopezMOB10,Galvan-Lopez2011}, and research is on going into finding and developing approaches to improve their overall performance including promoting neutrality in deceptive landscapes~\cite{10.1007/978-3-540-73482-6_9,DBLP:journals/tec/PoliL12}, dynamic fitness cases~\cite{DBLP:conf/gecco/LopezVST17Poster,DBLP:conf/ae/LopezVST17}, to mention a few examples.


An area of research which has proven popular in advancing the field of GP has been semantics. There have been a number of definitions for semantics in the past but broadly speaking semantics can be defined as the behavioural output of a program when executed using a set of inputs.

\begin{description}
\item {\textit{Semantic Similarity-based Crossover (SSC).}} This method was first proposed by Uy et al.~\cite{Uy2011} in the context of single-optimisation GP. This method uses a computationally expensive procedure by applying crossover between two parents multiple times using semantic diversity as a criteria in the selection process.

\item {\textit{Semantic-based Distance as an additional criteriOn (SDO).}} This is a method which originates as an improvement to the crowding distance operation as proposed by Galv\'an et al.~\cite{DBLP:conf/gecco/GalvanS19}. This method uses a reference point or pivot from the sparsely populated region of the search point and computes the semantic distance between this pivot and every individual. This distance is optimised as an additional criterion in Evolutionary Multiobjective Optimisation. 

\item {\textit{Pivot Semantic-based Distance as an additional criteriOn (PSDO).}} It is a method which is a variation of SDO but instead prefers solutions that are semantically similar to the pivot.
\end{description}

The goal of this paper are threefold: (i) to show how it is possible to \textit{naturally} incorporate semantics in MOGP leading to a better performance, (ii) to demonstrate the robustness of the proposed method by using two different forms of computing semantic distance, used in SDO and PSDO, leading to similar results by these two methods, and (iii) to show how a widely successful form of semantics used in single-objective GP does not necessarily yield good results in MOGP.

The structure of this paper is as follows. Section \ref{sec:rel} includes
the related work. Section \ref{sec:back} outlines the background in semantics and covers multi-objective genetic programming techniques. Section \ref{sec:sdo} presents the semantic-based approaches proposed in this work. Section \ref{sec:exp} explains the experimental
setup used and Section \ref{sec:res} shows and discusses the results obtained by the various semantic and canonical methods used in this study. The final section offers concluding remarks.\\



\section{Related Work}
\label{sec:rel}

\subsection{Semantics in Genetic Programming}

Scientific studies of semantics in GP have increased dramatically over the last years given that it has been consistently reported to be beneficial in GP search, ranging from the study of geometric operators~\cite{DBLP:conf/ppsn/MoraglioKJ12}, including the analysis of indirect semantics~\cite{Galvan-Lopez2016,Uy2011}. We discuss next the most relevant works to the research discussed in this paper. 

Even though researchers have proposed a variety of \sloppy{mechanisms} to use the semantics of GP programs to guide a search, it is commonly accepted that semantics refers to the output of a GP program once it is executed on a data set (also known as fitness cases in the specialised GP literature). The work conducted by McPhee et al.~\cite{McPhee:2008:SBB:1792694.1792707} paved the way for the proliferation of indirect semantics works. In their research, the authors studied the semantics of subtrees and the semantics of context (the remainder of a tree after the removal of a subtree). In their studies, the authors pointed out how a high proportion of individuals created by the widely used 90-10 crossover operator (i.e., 90\%-10\% internal-external node selection policy) are semantically equivalent. That is, the  crossover operator does not have any useful impact on the semantic GP space, which in consequence leads to a lack of performance increase as evolution continues.


Uy et al.~\cite{Nguyen:2009:SAC:1533497.1533524} proposed four different forms of applying semantic crossover operators on real-valued problems (e.g., symbolic regression problems). To this end, the authors measured the semantic equivalence of two expressions by measuring them against a random set of points sampled from the domain. If the resulting outputs of these two expressions were close to each other, subject to a threshold value called semantic sensitivity, these expressions were regarded as semantically equivalent. In their experimental design, the authors proposed four scenarios. In their first two scenarios, Uy et al. focused their attention on the semantics of subtrees. More specifically, for Scenario I, they tried to encourage semantic diversity by repeating crossover for a number of trials if two subtrees were semantically equivalent. Scenario II explored the opposite idea of Scenario I. For the last two scenarios, the authors focused their attention on the entire trees. That is, for Scenario III Uy et al. checked if offspring and parents were semantically equivalent. If so, the parents were transmitted into the following generation and the offspring were discarded. The authors explored the opposite of this idea in Scenario IV (children semantically different from their parents). They showed, for a number of symbolic regression problems, that Scenario I produced better results compared to the other three scenarios.

The major drawback with the Uy et al.~\cite{Nguyen:2009:SAC:1533497.1533524} approach is that it can be computational expensive, since it relies on a trial mechanism that attempts to find semantically different individuals via the execution of the crossover operator multiple times. To overcome this limitation, Galv\'an et al.~\cite{6557931} proposed a cost-effective mechanism based on the tournament selection operator to promote semantic diversity. More specifically, the tournament selection of the first parent is done as usual. That is, the fittest individual is chosen from a pool of individuals randomly picked from the population. The second parent is chosen from a pool of individuals that are semantically different from the first parent and it is also the fittest individual. If there is no individual semantically different from the first parent, then the tournament selection of the second parent is performed as usual. The proposed approach resulted in similar, and in some cases better, results compared to those reported by Uy et al.~\cite{Nguyen:2009:SAC:1533497.1533524,Uy2011} without the need of a trial and error (expensive) mechanism.



More recently, Forstenlechner et al.~\cite{Forstenlechner:2018:TES:3205455.3205592}   proposed two semantic operators, Effective Semantic (ES) Crossover for Program Synthesis and ES Mutation for Program Synthesis. Program synthesis operates on a range of different data types as opposed to those that work in a single data type such as real-valued cases~\cite{Uy2010} and Boolean values~\cite{6557931}. The main elements considered by the authors were the metrics used to determine semantic similarity, named partial change, used in the first instance,  and any change, used only if the first failed to be satisfied to avoid using standard crossover.  In their results, the authors reported that a semantic-based GP system achieved better results in 4 out of 8 problems used in their studies.

\subsection{Multi-Objective Genetic Programming}

In a multi-objective optimisation (MO) problem, one optimises with respect to multiple goals or objective functions. Thus, the task of the algorithm is to find acceptable solutions by considering all the criteria simultaneously. This can be achieved in various ways, where keeping the objectives separate is the most common. This form keeps the objectives separate and uses the notion of \textit{Pareto dominance}. In this way, Evolutionary MO (EMO)~\cite{1597059,CoelloCoello1999,Deb:2001:MOU:559152} offers an elegant solution to the problem of optimising two or more conflicting objectives. The aim of EMO is to simultaneously evolve a set of the best tradeoff solutions along the objectives in a single run.  EMO is one of the most active research areas in EAs thanks to its wide applicability as well as the impressive results achieved by these techniques~\cite{1597059,CoelloCoello1999,Deb:2001:MOU:559152}. 


MOGP has been used to classify highly unbalanced binary data~\cite{6198882,Galvan-Lopez2016}. To do so, the authors treated each objective (class) `separately' using EMO approaches~\cite{Deb02afast,Zitzler01spea2:improving}. Bhowan et al.~\cite{6198882} and Galv\'an et al.~\cite{DBLP:conf/gecco/GalvanS19,Galvan-Lopez2016,Galvan_MICAI_2016} showed, independently, how MOGP was able to achieve high accuracy in classifying binary data in conflicting learning objectives (i.e., normally a high accuracy of one class results in lower accuracy on the other).

\section{Background}
\label{sec:back}

\subsection{Semantics}


Pawlak et al. \cite{6808504} gave a formal definition for program semantics. Let $p \in P$ be a program from a given programming language $P$. The program $p$ will produce a specific output $p(in)$ where input $in \in I$. The set of inputs $I$ can be understood as being mapped to the set of outputs $O$ which can be defined as $p:I \rightarrow O$. 

\textbf{ Def 1.} \emph{Semantic mapping function is a function $s:P \rightarrow S$ mapping any program $p$ from $P$ to its semantic $s(p)$, where we can show the semantic equivalence of two programs};

\begin{equation}
s(p_1) = s(p_2) \iff  \forall\ in \in I: p_1(in) = p_2(in)
\end{equation}

This definition presents three important and intuitive properties for semantics:

\begin{enumerate}
\item Every program has only one semantic attributed to it.
\item Two or more programs may have the same semantics.
\item Programs which produce different outputs have different semantics.
\end{enumerate}

In Def. 1, we have not given a formal representation of semantics. In the following work semantics will be represented as a vector of output values which are executed by the program under consideration using an input set of data. For this representation of semantics we need to define semantics under the assumption of a finite set of fitness cases, where a fitness case is a pair comprised of a program input and its respective program output $I$ $\times$ $O$. This allows us to define the semantics of a program as follows:

\textbf{Def 2.} \emph{The semantics $s(p)$ of a program $p$ is the vector of
values from the output set $O$ obtained by computing $p$ on all inputs from the input set $I$}:

\begin{equation}
s(p) = [p(in_1), p(in_2), ... , p (in_l)]
\end{equation}

\noindent where $l = |I|$ is the size of the input set. Now that we have a formal definition in of semantics we can discuss the application of it in GP as elaborated in Section \ref{sec:sdo}.

\subsection{Pareto dominance}

In general terms a multi-objective problem seeks to find a solution that either maximizes or minimizes a number of objectives. In the case of maximization this can be represented mathematically as

\begin{equation}
max(f_1(x), f_2(x), ..., f_k(x))\ \ \ \ \ s.t. \ \ x \in X,
\label{eqn:mo}
\end{equation}

\noindent where X represents the feasible solution set, $f_i(x)$ represents the $i^{th}$ objective function for the feasible solution x and k $\geq$ 2. Typically there will not exist a unique solution that will maximize all objective functions. 
A candidate solution is Pareto dominant if its fitness is better or equal for all objectives and is strictly preferred by at least one in the search space. This can be formally represented  by

\begin{equation}
S_i \succ S_j \leftrightarrow \forall _m [ ( S_i )_m \ge  (S_j)_m] \land \exists k [ ( S_i )_k >  (S_j)_k ]
\label{eqn:pd}
\end{equation}

\noindent where $(S_i)_m$ is the $i^{th}$ solution for objective $m$ and $S_i \succ S_j$ denotes that solution $i$ is non-dominated by solution $j$. A candidate solution is considered Pareto optimal if is not dominated by any other candidate solution. In other words, if none of the objectives for a candidate solution can be improved without degrading at least one of the other objectives it can be considered Pareto optimal. 
For multi-objective problems there may exist a number of non-dominated solutions.
The set of non-dominated candidate solutions for an MO problem is referred to as the first Pareto frontier when represented in objective space. In practise it is not always possible to do an exhaustive search for the true Pareto optimal set and as such this is something we seek to approximate instead.
Pareto dominance relation is an integral part of MOEAs and has allowed practitioners and researchers to form important metrics in the selection process of these algorithms. Two such metrics are dominance rank and dominance count. Dominance rank is used as a fitness measure and calculates how may other solutions a candidate solution is dominated by. The lower the dominance rank the better with the lowest dominance rank of 0, i.e a solution that is not dominated by any other solution. Dominance rank can be expressed mathematically as seen in Equation \ref{eqn:dom_rank};

\begin{equation}
D_{rank}(S_i) = |\{j|j \in Pop \land S_j \succ S_i \}|
\label{eqn:dom_rank}
\end{equation}

\noindent where $|.|$ represents the cardinality of the set. This criteria is utilised in NSGA-II. Dominance count calculates how many individuals a candidate solution dominates [1]. The higher the dominance count the better.
Dominance rank and Dominance count are both used in SPEA-2. 

\begin{equation}
D_{count}(S_i) = |\{j|j \in Pop \land S_i \succ S_j \}|
\end{equation}

\noindent Leading on from the previous equation we can get a measure of raw fitness $R(i)$, by summing the fitness of all individuals such that [4];

\begin{equation}
R(i) = \sum_{j \in P_t + \bar{P}_t,\ j \succ i } D_{count}(S_j)
\end{equation}

\noindent A $R(i)$ value of 0 corresponds to an individual which is non-dominated and a $R(i)$ value that is high  corresponds to an individual which is dominated by many other individuals, as such the raw fitness should be minimized.

  \begin{table*}
\caption{Binary imbalanced classification datasets used in our research}
\centering
\resizebox{0.90\textwidth}{!}{ 
\begin{tabular}{llrrrrrr}
\hline
Data set & Classes             & \multicolumn{3}{c}{Number of examples} & Imb. & \multicolumn{2}{c}{Features} \\
         & Positive/Negative (Brief description) &  Total   & Positive    &  Negative     & Ratio  & No. & Type \\ \hline
Ion    & Good/bad (ionsphere radar signal)     & 351  & 126 (35.8\%) & 225 (64.2\%)       &1:3     &34  & Real  \\
Spect  & Abnormal/normal (cardiac tom. scan)   & 267  & 55 (20.6\%)  &212 (79.4\%)        &1:4     & 22  & Binary  \\
Yeast$_1$& mit/other (protein sequence)         &1482   & 244 (16.5\%) & 1238 (83.5\%)     &1:6     & 8 & Real  \\
Yeast$_2$& me3/other (protein sequence)         & 1482  & 163 (10.9\%) & 1319 (89.1\%)     &1:9     & 8 & Real  \\
Abal$_1$ & 9/18 (biology of abalone)        & 731   & 42 (5.75\%) & 689 (94.25\%)     & 1:17       &  8 & Real \\
Abal$_2$ & 19/other (biology of abalone)     & 4177   & 32 (0.77\%) & 4145 (99.23\%)   &  1:130      &  8 & Real \\
\hline
\end{tabular}
}
\label{tab:datasets}
\end{table*}

\subsection{Crowding distance}
\label{sec:sub:cd}

Solutions are ranked relative to each other according to a metric known as the crowding distance. The crowding distance is used to compare any pair of solutions in search space and is used in NSGA-II and SPEA2 as Pareto Dominance alone only acts as a partial order of the solutions. The crowding distance calculation is comprised of three parts;

\begin{itemize}
	\item Initialize the distance $d$ to zero.
	\item Set the boundary solutions to $\infty$. These solutions are always selected due to this constraint.
	\item Calculate the average distance differences for an individual against its two nearest neighbours using the Manhattan distance, shown in Equation~\ref{eqn:cd}:
	\begin{equation}
	d = d + \frac{|f^{(k)}_{r+1} -  f^{(k)}_{r-1}|}{|f^{(k)}_{max} -  f^{(k)}_{min}|}
	\label{eqn:cd}
	\end{equation}
\end{itemize}
	
\noindent where $k$ denotes the objective in question and r is the index for the current individual, where $r+1$ and $r-1$ reference its two nearest neighbours. Solutions with the highest crowding distance are considered better solutions, in other words the algorithm preferences localities along the Pareto front which are more sparsely populated with solutions than those which are more dense. In this manner the crowding distance resolves which solutions to retain when programs produce very similar fitness values.

\section{Semantic-based methods}
\label{sec:sdo}

\subsection{Semantic Similarity-based Crossover MOGP}
To incorporate semantics in a MOGP paradigm, we first use the Semantic Similarity-based Crossover (SSC) originally proposed by Uy et al.~\cite{Uy2011} which, to the best of our knowledge, has been exclusively used in single-objective GP.

To use SSC in single-objective GP a semantic distance must be computed first. Using Def. 2, this distance is obtained by computing the average of the absolute difference of values for every $in \in I$ between parent and offspring. If the distance value lies within a range, defined by one or two threshold values, then crossover is used to generate offspring. Because this condition may be hard to satisfy, the authors tried to encourage semantic diversity by repeatedly applying crossover up to 20 times. If after this, the condition is not satisfied, then crossover is executed as usual.


SSC made a notable impact in GP, showing, for the first time, how semantic diversity can be promoted in continuous search spaces, with several subsequent
papers following along this line~\cite{DBLP:conf/gecco/GalvanS19,6557931,Krawiec:GPEM2013}. We incorporate SSC in NSGA-II and SPEA2. However, the performance increase reported when adopting SSC in single-objective optimisation is not observed in MOGP, as discussed in Section~\ref{sec:res}. 

\subsection{Semantic Distance as an additional criteriOn}
\label{sec:sub:sdo}

In this method the crowding distance as discussed in Section \ref{sec:sub:cd} is replaced by a semantic-based crowding distance. A pivot $p$ is selected as the individual in the first Pareto front which is furthest away from all other individuals $v$ in that front. This distance is calculated using the crowding distance as discussed previously. Once we have the pivot, we can compute the semantic differences of the pivot against all the other individuals in the population. Upper and lower semantic similarity bounds, denoted as UBSS and LBSS, respectively, are used to promote semantic diversity within a range as shown in Equation~\ref{eqn:sd} or via a single bound as like in Equation~\ref{eqn:sd2}.

\begin{equation}
d(p, v_j) = \sum_{i=1}^{l} 1 \ if\ LBSS \leq |p(in_i) - v_j(in_i)| \leq UBSS
\label{eqn:sd}
\end{equation}

\begin{equation}
d(p, v_j) = \sum_{i=1}^{l} 1 \ if\  |p(in_i) - v_j(in_i)| \geq UBSS
\label{eqn:sd2}
\end{equation}

\noindent The semantic-based crowding distance can also be used as an additional criterion to optimise. With the majority and minority class (these benchmark problems are discussed in detail in the following section) serving as the first two objectives to optimize semantic distance is also treated a third objective.

\subsection{Pivot Similarity Semantic-based Distance as and additional criteriOn}
\label{sec:psdo}

In Section \ref{sec:sub:sdo} we discussed SDO which calculates the semantic distance between each individual and a pivot. To this effect Equations \ref{eqn:sd} and \ref{eqn:sd2} were used to determine if the semantic difference between a pivot and the individuals fall within a predefined range. This calculation naturally preferences programs which are semantically dissimilar to the pivot. However, given that the pivot is picked from the most sparse region of the search space this individual ought to be the most diverse as such an update a proposed update in the distance calculation is given in Equations~\ref{eqn:sdto} and \ref{eqn:sdto2}, where the summed values is now taken away from the number of fitness cases. This method is referred to as Pivot Similarity Semantic-based Distance as an additional criteriOn (PSDO) and uses Equations~\ref{eqn:sdto} and \ref{eqn:sdto2} in lieu of distances shown in Equations~\ref{eqn:sd} and \ref{eqn:sd2}.

\begin{equation}
d(p, v_j) = l - \sum_{i=1}^{l} 1 \ if\ LBSS \leq |p(in_i) - v_j(in_i)| \leq UBSS
\label{eqn:sdto}
\end{equation}

\begin{equation}
d(p, v_j) = l - \sum_{i=1}^{l} 1 \ if\  |p(in_i) - v_j(in_i)| \geq UBSS
\label{eqn:sdto2}
\end{equation}\\

\section{Experimental Setup}
\label{sec:exp}


The data sets originate from the well-known UCI Machine Learning repository \cite{Dua:2019}. The characteristics of these data sets vary greatly, with varying number of features and imbalance ratios. A brief description of each of these data sets is given in Table \ref{tab:datasets}, including the nature of their feature type, their imbalance ratios and the number of features. All data sets were split 50/50 with half of the entries being attributed to the training set and the other half for the test set. The same class imbalance ratio is kept between the training and test set. All the results reported in this work are based on the latter. 
The function set is the list of arithmetic operators used by the GP and are assigned at the non-terminal nodes, these were selected as $ \Re = \{+, -, *, \%\} $, where \% denotes the protected division operator.

\begin{table}
\centering
\caption{Summary of parameters}
\resizebox{0.85\columnwidth}{!}{ 
\small\begin{tabular}{|l|r|} \hline 
\emph{Parameter} &
\emph{Value} \\ \hline \hline
Population Size & 500 \\ \hline
Generations & 50 \\ \hline
Type of Crossover & 90\% internal nodes, 10\% leaves  \\ \hline
Crossover Rate  & 0.60  \\ \hline
Type of Mutation & Subtree \\ \hline
Mutation Rate & 0.40 \\ \hline
Selection & Tournament (size = 7) \\ \hline
Initialisation Method & Ramped half-and-half \\ \hline
Initialisation Depths: & \\ 
\hspace{.3cm}Initial Depth & 1 (Root = 0)\\ 
\hspace{.3cm}Final Depth & 5 \\ \hline
Maximum Length & 800 \\ \hline
Maximum Final Depth & 8\\ \hline
Independent Runs & 50 \\ \hline
Semantic Thresholds &  UBSS = \{0.25, 0.5, 0.75, 1.0\} \\
 &  LBSS = \{-, 0.001, 0.01, 0.1\} \\ \hline

\end{tabular}
}
\label{tab:parameters}
\end{table}



A common metric used in determining fitness for binary classification problems, as the ones used in this work, is to use classification accuracy; where $ACC = \frac{TP + TN}{TP + TN + FP + FN}$. However with imbalanced data sets using this accuracy measure will tend to bias towards the majority class. As such it is better to treat the minority and majority as two separate objectives where the goal is to maximize the number of correctly classified cases. This can be done using the true positive rate $TPR = \frac{TP}{TP + FN}$ and true negative $TNR = \frac{TN}{TN + FP}$  \cite{6198882}.  Table \ref{tab:parameters} gives an overview of the parameters used in our work. Of note are the UBSS and LBSS settings (bottom of Table~\ref{tab:parameters}). These allow for a possible combination of 16 settings for the semantic thresholds. Each setting is run for 50 independent runs which results in 4800 independent runs for each semantic-based method tested. In short, when we consider  all of the semantic-based methods in addition to the canonical methods for both MOGP frameworks over 29,400 independent runs were conducted. 
All experiments were run on Kay a supercomputer with 336 nodes operated by Irish Centre for High-End Computing.


\begin{table}
\caption{Average hypervolume ($\pm$ std. deviation) and last run Pareto Front for NSGA-II and SPEA2 for 50 independent runs.}
\centering
\resizebox{1\columnwidth}{!}{
\begin{tabular}{ccccc}\hline
\multirow{3}{*}{Dataset} & \multicolumn{2}{c}{NSGA-II}        & \multicolumn{2}{c}{SPEA2} \\
           & \multicolumn{2}{c}{Hypervolume}       & \multicolumn{2}{c}{Hypervolume} \\
    &  Average & PO Front     & Average & PO Front     \\ \hline

Ion &  0.766 $\pm$ 0.114 & 0.938 &  0.786 $\pm$ 0.094  & 0.948 \\

 Spect &  0.534 $\pm$ 0.024  & 0.647 & 0.544 $\pm$ 0.032 & 0.659  \\

Yeast$_1$ & 0.838 $\pm$ 0.011  & 0.876  & 0.838 $\pm$ 0.008 & 0.877 \\

 Yeast$_2$ & 0.950 $\pm$ 0.009  & 0.976  & 0.946 $\pm$ 0.015 &  0.978 \\

 Abal$_1$ & 0.847 $\pm$ 0.058 &  0.961 &  0.832 $\pm$ 0.078 &  0.960 \\

 Abal$_2$ &0.576 $\pm$ 0.122 & 0.842  & 0.544 $\pm$ 0.147 & 0.834 \\
\hline
\end{tabular}
}
\label{tab:hyperarea:nsgaii:spea2}
\end{table}

\section{Results and Analysis}
\label{sec:res}

In order to  compare the  approaches used in this work, we use the hypervolume~\cite{Coello:2006:EAS:1215640} of the evolved Pareto approximations as a performance measure. For bi-objective problems, as the binary highly unbalanced classifications problems used in this work are, the hypervolume of a set of points in objective space, using reference point $(0,0)$, is easily computed as the sum of the areas of all trapezoids fitted under each point. The reference point $(0,0)$ is required for the hypervolume calculation and as we wish to maximize each of the objectives we select this point as the lowest value that either the $TPR$ or $TNR$ can take. The hypervolume was chosen because it is the one of the most widely used performance indicator in the EMO literature. We also computed the accumulated Pareto-optimal (PO) front with respect to 50 runs: the set of non-dominated solutions after merging all 50 Pareto-approximated fronts. 

Tables~\ref{tab:hyperarea:nsgaii:spea2} and~\ref{tab:nsgaii:semantics}  report, for each problem defined in Table~\ref{tab:datasets} both, the average hypervolume over 50 runs and also the hypervolume of the accumulated PO front with respect to all 50 runs. To obtain a statistically sound conclusion, a series of  Wilcoxon rank-sum tests were run on the average hypervolume results. To account for the problem of  multiple comparisons that arose from testing the canonical method 16 times for each data set, a Bonferroni correction $\frac{\alpha}{m} = 3.125 \times 10^{-3}$ was used where $\alpha$ = 0.05. These statistically significant differences are highlighted in boldface in Table~\ref{tab:nsgaii:semantics} (this shown in the appendix). Moreover, in this table, the symbols ``+'', ``--'', and ``='' indicate that the results of a given semantic-based approach are significantly better, worse, or equal, respectively, than those found by the canonical NSGA-II (Table~\ref{tab:hyperarea:nsgaii:spea2}), all the above based on the Wilcoxon rank-sum test.

Let us first focus our attention on the average hypervolume and PO front results yield by NSGA-II and SPEA2 shown in Table~\ref{tab:hyperarea:nsgaii:spea2}. Such observed differences between them are not statistically significant in any of the dataset used. We can now compare the semantic-based methods, introduced in Section~\ref{sec:sdo}, against their corresponding canonical EMO algorithms. Due to the significant amount of results obtained and shown in the appendix, we summarised our findings in a series of payoff matrices to show which strategies have significant better values or `wins' \textit{vs.} the other methods. These payoff matrices are shown in Tables~\ref{payoff:nsgaii} and~\ref{payoff:spea2} for semantic-based methods using NSGA-II and SPEA2, respectively.

\begin{table}[tb]
\centering
  \caption{Payoff tables for canonical NSGA-II, NSGA-II SDO, NSGA-II PSDO and NSGA-II SSC for each of the 6 data sets}
\resizebox{1\columnwidth}{!}{
    \begin{tabularx}{0.30\textwidth}{cc|c|c|c|c|c|}

      & \multicolumn{1}{c}{} &
      \multicolumn{4}{c}{\textbf{Ion}}  \\
       \\
      & \multicolumn{1}{c}{} &

      \multicolumn{1}{L}{\mbox{CAN.}}  & \multicolumn{1}{Y}{SDO} & \multicolumn{1}{Y}{PSDO}  & \multicolumn{1}{Y}{SSC}
      
      \\\cline{3-6}

      \multirow{4}* & CAN. & - & 0 \cellcolor{YYYY}  & 0 \cellcolor{YYYY}  &  0 \cellcolor{YYYY} 
      
      \\\cline{3-6}
      
      & SDO & \textcolor{white}{16} \cellcolor{RRRR}  & - & 2 \cellcolor{OOOY} & \textcolor{white}{15} \cellcolor{RRRR}
      \\\cline{3-6}
            & PSDO & \textcolor{white}{13} \cellcolor{RRRO} & {3} \cellcolor{OOYY} & - & \textcolor{white}{14} \cellcolor{RRRO}
      \\
      \cline{3-6}
            & SSC & 0 \cellcolor{YYYY} & 0 \cellcolor{YYYY}  & 0 \cellcolor{YYYY} & -
      \\\cline{3-6}

    \end{tabularx}
    \begin{tabularx}{0.30\textwidth}{cc|c|c|c|c|c|}

      & \multicolumn{1}{c}{} &
      \multicolumn{4}{c}{\textbf{Spect}}  \\
       \\
      & \multicolumn{1}{c}{} &

      \multicolumn{1}{L}{\mbox{CAN.}}  & \multicolumn{1}{Y}{SDO} & \multicolumn{1}{Y}{PSDO}  & \multicolumn{1}{Y}{SSC}
      
      \\\cline{3-6}

      \multirow{4}* & CAN. & - & 0 \cellcolor{YYYY} & 0 \cellcolor{YYYY} &  0 \cellcolor{YYYY} 
      \\\cline{3-6}
      & SDO & \textcolor{white}{16} \cellcolor{RRRR} & - & \textcolor{white}{8} \cellcolor{OOOO} & \textcolor{white}{16} \cellcolor{RRRR}
      \\\cline{3-6}
            & PSDO & \textcolor{white}{16} \cellcolor{RRRR} & 3 \cellcolor{OOYY} & - & \textcolor{white}{15} \cellcolor{RRRR}
      \\
      \cline{3-6}
            & SSC & 0 \cellcolor{YYYY} & 0 \cellcolor{YYYY}  & 0 \cellcolor{YYYY} & -
      \\\cline{3-6}

    \end{tabularx}
    }

\centering
\resizebox{1.0\columnwidth}{!}{
    \begin{tabularx}{0.30\textwidth}{cc|c|c|c|c|c|}

      & \multicolumn{1}{c}{} &
      \multicolumn{4}{c}{\textbf{Yeast$_1$}}  \\
       \\
      & \multicolumn{1}{c}{} &

      \multicolumn{1}{L}{\mbox{CAN.}} & \multicolumn{1}{Y}{SDO} & \multicolumn{1}{Y}{PSDO}  & \multicolumn{1}{Y}{SSC}
      
      \\\cline{3-6}

      \multirow{4}* & CAN. & - & 0 \cellcolor{YYYY}  & 0 \cellcolor{YYYY}  &  {0} \cellcolor{YYYY} 
      
      \\\cline{3-6}
      
      & SDO & \textcolor{white}{16} \cellcolor{RRRR}  & - & 2 \cellcolor{OYYY} & \textcolor{white}{16} \cellcolor{RRRR}
      \\\cline{3-6}
            & PSDO & \textcolor{white}{16} \cellcolor{RRRR} & {1} \cellcolor{OYYY} & - & \textcolor{white}{16} \cellcolor{RRRR} 
      \\
      \cline{3-6}
            & SSC & 0 \cellcolor{YYYY} & 0 \cellcolor{YYYY}  & 0 \cellcolor{YYYY} & -
      \\\cline{3-6}

    \end{tabularx}
    \begin{tabularx}{0.30\textwidth}{cc|c|c|c|c|c|}

      & \multicolumn{1}{c}{} &
      \multicolumn{4}{c}{\textbf{Yeast$_2$}}  \\
       \\
      & \multicolumn{1}{c}{} &

      \multicolumn{1}{L}{\mbox{CAN.}}  & \multicolumn{1}{Y}{SDO} & \multicolumn{1}{Y}{PSDO}  & \multicolumn{1}{Y}{SSC}
      
      \\\cline{3-6}

      \multirow{4}* & CAN. & - & 0 \cellcolor{YYYY} & 0 \cellcolor{YYYY} &  0 \cellcolor{YYYY}
      \\\cline{3-6}
      & SDO & \textcolor{white}{16} \cellcolor{RRRR} & - & 2 \cellcolor{OYYY} & \textcolor{white}{16} \cellcolor{RRRR}
      \\\cline{3-6}
            & PSDO & \textcolor{white}{16} \cellcolor{RRRR} & {4} \cellcolor{OOOY} & - & \textcolor{white}{16} \cellcolor{RRRR}
      \\
      \cline{3-6}
             & SSC & 0 \cellcolor{YYYY} & 0 \cellcolor{YYYY}  & 0 \cellcolor{YYYY} & -
      \\\cline{3-6}

    \end{tabularx}
    }
  

\centering
\resizebox{1.0\columnwidth}{!}{
    \begin{tabularx}{0.30\textwidth}{cc|c|c|c|c|c|}

      & \multicolumn{1}{c}{} &
      \multicolumn{4}{c}{\textbf{Abal$_1$}}  \\
       \\
      & \multicolumn{1}{c}{} &

      \multicolumn{1}{L}{\mbox{CAN.}}  & \multicolumn{1}{Y}{SDO} & \multicolumn{1}{Y}{PSDO}  & \multicolumn{1}{Y}{SSC}
      
      \\\cline{3-6}

      \multirow{4}* & CAN. & - & 0 \cellcolor{YYYY} & 0 \cellcolor{YYYY} & 0 \cellcolor{YYYY}
      
      \\\cline{3-6}
      
      & SDO &  \textcolor{white}{12}  \cellcolor{RROO} & - &  \textcolor{white}{10} \cellcolor{RROO} & \textcolor{white}{12} \cellcolor{RROO}
      \\\cline{3-6}
            & PSDO &  \textcolor{white}{6} \cellcolor{OOOY} & {1} \cellcolor{OYYY} & - & \textcolor{white}{7} \cellcolor{OOOO}
      \\
      \cline{3-6}
            & SSC & 0 \cellcolor{YYYY} & 0 \cellcolor{YYYY}  & 0 \cellcolor{YYYY} & -
      \\\cline{3-6}

    \end{tabularx}
    \begin{tabularx}{0.30\textwidth}{cc|c|c|c|c|c|}

      & \multicolumn{1}{c}{} &
      \multicolumn{4}{c}{\textbf{Abal$_2$}}  \\
       \\
      & \multicolumn{1}{c}{} &

      \multicolumn{1}{L}{\mbox{CAN.}}  & \multicolumn{1}{Y}{SDO} & \multicolumn{1}{Y}{PSDO}  & \multicolumn{1}{Y}{SSC}
      
      \\\cline{3-6}

      \multirow{4}* & CAN. & - & 0 \cellcolor{YYYY} & 0 \cellcolor{YYYY} &  {2} \cellcolor{OYYY}
      \\\cline{3-6}
      & SDO & \textcolor{white}{12} \cellcolor{RROO} & - & 1 \cellcolor{OYYY}  & \textcolor{white}{15} \cellcolor{RRRR}
      \\\cline{3-6}
            & PSDO & \textcolor{white}{16} \cellcolor{RRRR} & \textcolor{white}{7} \cellcolor{OOOO} & - & \textcolor{white}{16} \cellcolor{RRRR}
      \\
      \cline{3-6}
            & SSC & 0 \cellcolor{YYYY} & 0 \cellcolor{YYYY}  & 0 \cellcolor{YYYY} & -
      \\\cline{3-6}

    \end{tabularx}
    }
  \\
  \label{payoff:nsgaii}
  \end{table}

The table can be read as follows: the strategies of the row index are compared against the strategies of the column index and for each LBSS and UBSS setting which is significantly better for the column strategy counts as one `win' towards the count. For example SDO \textit{vs.} NSGA-II for the Ion data set is significantly better for all settings of LBSS and UBSS and as such has 16 `wins' overall (top left-hand side of Table~\ref{payoff:nsgaii}). The tables have been colour coded as such with solid black denoting that the strategy is the best overall in terms of the number of `wins' and light grey being the worst overall. We do not show all the results for SPEA2 and the semantic-based methods, as we did for NSGA-II, shown in Table~\ref{tab:nsgaii:semantics}, due to space constraints, but the payoff table indicates how SDO and PSDO consistently yield better results compared to canonical SPEA2. 

The two methods that use semantic distance as a criteria strategies, SDO and PSDO, outperformed their respective canonical counterparts with the exception of Abal$_1$ which had a large number of settings that produced no significant difference in the hypervolume averages for canonical NSGA-II. Additionally for Abal$_1$ NSGA-II SDO had the greatest number of wins over other strategies and results for NSGA-II PSDO were comparatively mixed.

If we consider just SDO \textit{vs.} PSDO, results tend to be mixed. NSGA-II PSDO produced more wins for certain data sets like Yeast$_2$ and Abal$_2$ when compared with NSGA-II SDO but under performed for Spect and Abal$_1$. Typically when NSGA-II PSDO out performed against NSGA-II SDO, it was for settings that were held constant. For instance, three of the wins associated with Spect were a result of keeping UBSS constant at 1.0 with LBSS values of (0.001, 0.01, 0.1). NSGA-II PSDO performed significantly worse most often when LBSS was undefined except for Abal$_1$ where the results were reversed, i.e when LBSS was undefined PSDO performed as good or significantly better but was significantly worse for 10 of the other LBSS settings. There was little or no significant difference when comparing SPEA2 SDO and SPEA2 PSDO strategies with only SPEA2 SDO vs SPEA2 PSDO in Yeast$_1$ producing 1 `Win'. From this we can conclude that both methods, SDO and PSDO, that treat semantic distance as an additional criterion to be optimised perform similar and yield consistently better results to SSC. This latter method is based on the notion of single-objective GP adapted to MOGP, showing that the benefits obversed in SOGP are depleted in MOGP.

%




\begin{table}[tb]
\centering
    \caption{Payoff tables for canonical SPEA2, SPEA2 SDO, SPEA2 PSDO and SPEA2 SSC for each of the 6 data sets.}

\resizebox{1.0\columnwidth}{!}{
    \begin{tabularx}{0.30\textwidth}{cc|c|c|c|c|c|}

      & \multicolumn{1}{c}{} &
      \multicolumn{4}{c}{\textbf{Ion}}  \\
       \\
      & \multicolumn{1}{c}{} &

      \multicolumn{1}{Y}{CAN.}  & \multicolumn{1}{Y}{SDO} & \multicolumn{1}{Y}{PSDO}  & \multicolumn{1}{Y}{SSC}
      
      \\\cline{3-6}

      \multirow{4}* & CAN. & - & 0 \cellcolor{YYYY} & 0 \cellcolor{YYYY} &  0 \cellcolor{YYYY}
      
      \\\cline{3-6}
      
      & SDO & \textcolor{white}{16} \cellcolor{RRRR} & - & 0 \cellcolor{YYYY} & \textcolor{white}{16} \cellcolor{RRRR}
      \\\cline{3-6}
            & PSDO & \textcolor{white}{16} \cellcolor{RRRR} & 0 \cellcolor{YYYY} & - & \textcolor{white}{16} \cellcolor{RRRR}
      \\
      \cline{3-6}
            & SSC & 0 \cellcolor{YYYY} & 0 \cellcolor{YYYY}  & 0 \cellcolor{YYYY} & -
      \\\cline{3-6}

    \end{tabularx}
    \begin{tabularx}{0.30\textwidth}{cc|c|c|c|c|c|}

      & \multicolumn{1}{c}{} &
      \multicolumn{4}{c}{\textbf{Spect}}  \\
       \\
      & \multicolumn{1}{c}{} &

      \multicolumn{1}{Y}{CAN.}  & \multicolumn{1}{Y}{SDO} & \multicolumn{1}{Y}{PSDO}  & \multicolumn{1}{Y}{SSC}
      
      \\\cline{3-6}

      \multirow{4}* & CAN. & - & 0 \cellcolor{YYYY} & 0 \cellcolor{YYYY} & 0 \cellcolor{YYYY}
      \\\cline{3-6}
      & SDO & \textcolor{white}{14} \cellcolor{RRRR} & - & 0 \cellcolor{YYYY} & \textcolor{white}{16} \cellcolor{RRRR}
      \\\cline{3-6}
            & PSDO & \textcolor{white}{16} \cellcolor{RRRR} & 0 \cellcolor{YYYY} & - & \textcolor{white}{16} \cellcolor{RRRR}
      \\
      \cline{3-6}
              & SSC & 0 \cellcolor{YYYY} & 0 \cellcolor{YYYY}  & 0 \cellcolor{YYYY} & -
      \\\cline{3-6}

    \end{tabularx}
    }

\centering
\resizebox{1.0\columnwidth}{!}{
    \begin{tabularx}{0.30\textwidth}{cc|c|c|c|c|c|}

      & \multicolumn{1}{c}{} &
      \multicolumn{4}{c}{\textbf{Yeast$_1$}}  \\
       \\
      & \multicolumn{1}{c}{} &

      \multicolumn{1}{Y}{CAN.}  & \multicolumn{1}{Y}{SDO} & \multicolumn{1}{Y}{PSDO}  & \multicolumn{1}{Y}{SSC}
      
      \\\cline{3-6}

      \multirow{4}* & CAN. & - & 0 \cellcolor{YYYY} & 0 \cellcolor{YYYY} & 0 \cellcolor{YYYY}
      
      \\\cline{3-6}
      
      & SDO & \textcolor{white}{16} \cellcolor{RRRR} & - & 1 \cellcolor{OYYY} & \textcolor{white}{16} \cellcolor{RRRR}
      \\\cline{3-6}
            & PSDO & \textcolor{white}{16} \cellcolor{RRRR} & 0 \cellcolor{YYYY} & - & \textcolor{white}{16} \cellcolor{RRRR}
      \\
      \cline{3-6}
            & SSC & 0 \cellcolor{YYYY} & 0 \cellcolor{YYYY}  & 0 \cellcolor{YYYY} & -
      \\\cline{3-6}

    \end{tabularx}
    \begin{tabularx}{0.30\textwidth}{cc|c|c|c|c|c|}

      & \multicolumn{1}{c}{} &
      \multicolumn{4}{c}{\textbf{Yeast$_2$}}  \\
       \\
      & \multicolumn{1}{c}{} &

      \multicolumn{1}{Y}{CAN.}  & \multicolumn{1}{Y}{SDO} & \multicolumn{1}{Y}{PSDO}  & \multicolumn{1}{Y}{SSC}
      
      \\\cline{3-6}

      \multirow{4}* & CAN. & - & 0 \cellcolor{YYYY} & 0 \cellcolor{YYYY} &  0 \cellcolor{YYYY}
      \\\cline{3-6}
      & SDO & \textcolor{white}{16} \cellcolor{RRRR} & - & 0 \cellcolor{YYYY} & \textcolor{white}{16} \cellcolor{RRRR}
      \\\cline{3-6}
            & PSDO & \textcolor{white}{16} \cellcolor{RRRR} & 0 \cellcolor{YYYY} & - & \textcolor{white}{16} \cellcolor{RRRR}
      \\
      \cline{3-6}
            & SSC & 0 \cellcolor{YYYY} & 0 \cellcolor{YYYY}  & 0 \cellcolor{YYYY} & -
      \\\cline{3-6}

    \end{tabularx}
    }

\centering
\resizebox{1.0\columnwidth}{!}{
    \begin{tabularx}{0.30\textwidth}{cc|c|c|c|c|c|}

      & \multicolumn{1}{c}{} &
      \multicolumn{4}{c}{\textbf{Abal$_1$}}  \\
       \\
      & \multicolumn{1}{c}{} &

      \multicolumn{1}{Y}{CAN.}  & \multicolumn{1}{Y}{SDO} & \multicolumn{1}{Y}{PSDO}  & \multicolumn{1}{Y}{SSC}
      
      \\\cline{3-6}

      \multirow{4}* & CAN. & - & 0\cellcolor{YYYY}  & 0  \cellcolor{YYYY} &  0 \cellcolor{YYYY}
      
      \\\cline{3-6}
      
      & SDO & \textcolor{white}{16} \cellcolor{RRRR} & - & O \cellcolor{YYYY} & \textcolor{white}{16} \cellcolor{RRRO}
      \\\cline{3-6}
            & PSDO & \textcolor{white}{13} \cellcolor{RRRO} & 0  \cellcolor{YYYY} & - & \textcolor{white}{13} \cellcolor{RRRO}
      \\
      \cline{3-6}
            & SSC & 0 \cellcolor{YYYY} & 0 \cellcolor{YYYY}  & 0 \cellcolor{YYYY} & -
      \\\cline{3-6}

    \end{tabularx}
    \begin{tabularx}{0.30\textwidth}{cc|c|c|c|c|c|}

      & \multicolumn{1}{c}{} &
      \multicolumn{4}{c}{\textbf{Abal$_2$}}  \\
       \\
      & \multicolumn{1}{c}{} &

      \multicolumn{1}{Y}{CAN.}  & \multicolumn{1}{Y}{SDO} & \multicolumn{1}{Y}{PSDO}  & \multicolumn{1}{Y}{SSC}
      
      \\\cline{3-6}

      \multirow{4}* & SPEA2 & - & 0 \cellcolor{YYYY} & 0 \cellcolor{YYYY} &  0 \cellcolor{YYYY} 
      \\\cline{3-6}
      & SDO & \textcolor{white}{15} \cellcolor{RRRR} & - & 0 \cellcolor{YYYY} & \textcolor{white}{16} \cellcolor{RRRR}
      \\\cline{3-6}
            & PSDO & \textcolor{white}{16} \cellcolor{RRRR} & 0 \cellcolor{YYYY} & - & \textcolor{white}{16} \cellcolor{RRRR}
      \\
      \cline{3-6}
            & SSC & 0 \cellcolor{YYYY} & 0 \cellcolor{YYYY}  & 0 \cellcolor{YYYY} & -
      \\\cline{3-6}

    \end{tabularx}
    }

    \label{payoff:spea2}
  \end{table}
  
  \bigskip
 

\section{Conclusions}

The analysis and experimentation conducted for this paper clearly show the benefits of incorporating semantics in MOGP. A key finding was that regardless of which method was used in calculating semantic distance as an additional  criterion to optimise, each method (SDO and PSDO) proved to be significantly better than canonical methods as well as SSC, a method continously reported to be beneficial in single-objective GP.

When the various semantic distance methods were analysed it was found these methods had a tendency to preference programs that were semantically very similar and also semantically very dissimilar relative to the pivot. The pivot represents an individual from the most sparse region of the search space and can be considered the most divergent relative to the other programs in the population, as such it makes sense that these programs would be preferenced. Additionally the more semantically diverse programs will produce programs with significantly different outputs and ought to also be preferenced.

As the semantic distance is not an objective that is in conflict with the majority and minority class, both the SDO and PSDO methods are versatile enough to satisfy both of these criteria. The results showed that both these methods performed significantly better than canonical methods as well as the semantic-based method of SSC. While it was observed that both semantically similar and semantically dissimilar individuals relative to the pivot are preferable, it has not been shown if the specific location of the pivot in search space has any bearing on which distance metric would produce more diverse solutions for a given generation. As such further analysis on the pivot and its location in objective space relative to other individuals, may be useful in determining if a strategy of switching between SDO and PSDO methods during runtime would be worthwhile. Future work would likely benefit further from semantic-based distance methods and their is room for for further exploration of how to calculate new distance metrics.


\section{acknowledgments}

\noindent This publication has emanated from research conducted with the financial support of Science Foundation Ireland under Grant number 18/CRT/6049. 
The authors wish to acknowledge the DJEI/DES/SFI/HEA Irish Centre for High-End Computing (ICHEC) for the provision of computational facilities and support. 

\bibliographystyle{abbrv}
\bibliography{semantics.bib}

\newpage 
\clearpage

\begin{samepage}
\onecolumn
\section*{Appendix}
\nopagebreak
\begin{table*}[!htbp]

  \caption{\scriptsize{Average hypervolume ($\pm$ std. deviation) and last run Pareto Front for NSGA-II SDO,  NSGA-II PSDO and NSGA-II SSC methods.}}
 \thispagestyle{empty}
  \centering
   \resizebox{0.733\textwidth}{!}{
  \begin{tabular}{cccccccccc}\hline

                             &                            & \multicolumn{8}{c}{Hypervolume}  \\
    &                            & \multicolumn{4}{c}{Average} & \multicolumn{4}{c}{PO Front}  \\
     &          & \multicolumn{4}{c}{UBSS} & \multicolumn{4}{c}{UBSS} \\ 
    &    LBSS              & 0.25 & 0.5& 0.75 & 1.0 & 0.25 & 0.5& 0.75 & 1.0 \\ \hline

    \multicolumn{10}{c}{\textsf{NSGA-II SDO}} \\ \hline
    \multirow{4}{*}{Ion} & -- & \textbf{0.860 $\pm$ 0.033}+ & \textbf{0.869 $\pm$ 0.037}+  & \textbf{\underline{0.869 $\pm$ 0.033}}+ & \textbf{0.845 $\pm$ 0.057}+ & {0.948} & {0.958} & {0.962} & {0.950}  \\
     & 0.001                & {\textbf{0.817 $\pm$ 0.087}}+ & \textbf{0.819 $\pm$ 0.104}+ & \textbf{0.857 $\pm$ 0.057}+ & \textbf{0.861 $\pm$ 0.047}+ & {0.942} & {0.957} & {0.954} & {0.958} \\ 
     & 0.01 & \textbf{0.825 $\pm$ 0.084}+ & \textbf{0.843 $\pm$ 0.073}+ & \textbf{0.861 $\pm$ 0.045}+ & \textbf{0.861 $\pm$ 0.038}+ & {0.946} & {0.956} & {0.957} & {0.944} \\
      & 0.1 & \textbf{0.846 $\pm$ 0.070}+ & \textbf{0.848 $\pm$ 0.068}+ &\textbf{0.844 $\pm$ 0.075}+ & \textbf{0.864 $\pm$ 0.044}+ & {0.950} & {0.956} & {0.953} & \underline{0.960} \\ \hline 
    
   \multirow{4}{*}{Spect} & -- & \textbf{0.591 $\pm$ 0.027}+  & \textbf{0.593 $\pm$ 0.025}+  & \textbf{0.594 $\pm$ 0.023}+ & \textbf{\underline{0.600 $\pm$ 0.019}}+ &  {0.684} & {0.679} & {0.689} & \underline{0.694} \\
                       & 0.001 & \textbf{0.562 $\pm$ 0.021}+ & {\textbf{0.558 $\pm$ 0.025}}+ & \textbf{0.561 $\pm$ 0.019}+  & \textbf{0.560 $\pm$ 0.016}+ & {0.668} & {0.653} & {0.660} & {0.644} \\ 
                       & 0.01  & \textbf{0.564 $\pm$ 0.025}+ & \textbf{0.560 $\pm$ 0.023}+ & \textbf{0.566 $\pm$ 0.024}+ & \textbf{0.559 $\pm$ 0.016}+ & {0.672} & {0.650} & {0.669} & {0.643} \\ 
                       & 0.1  & \textbf{0.563 $\pm$ 0.022}+ & \textbf{0.563 $\pm$ 0.024}+ & \textbf{0.567 $\pm$ 0.018}+ & \textbf{0.561 $\pm$ 0.024}+ & {0.664} & {0.658} & {0.655} & {0.658}  \\  \hline

  \multirow{4}{*}{Yeast$_1$} & -- & \textbf{0.850 $\pm$ 0.006}+ & \textbf{0.849 $\pm$ 0.008}+ & \textbf{0.849 $\pm$ 0.006}+ & \textbf{0.849 $\pm$ 0.006}+ & {0.881} & {0.881} & {0.882} & {0.881} \\ 
   & 0.001 & {\textbf{0.845 $\pm$ 0.007}}+ & \textbf{0.847 $\pm$ 0.006}+ & \textbf{0.848 $\pm$ 0.004}+ & \textbf{0.848 $\pm$ 0.005}+ & {0.879} & {0.882} & {0.879} & {0.880} \\ 
     & 0.01     & \textbf{0.848 $\pm$ 0.006}+  & \textbf{0.849 $\pm$ 0.005}+ & \textbf{0.848 $\pm$ 0.005}+ & \textbf{\underline{0.850 $\pm$ 0.005}}+ & {0.881} & {0.881} & {0.879} & {0.881} \\ 
      & 0.1     & \textbf{0.847 $\pm$ 0.005}+ & \textbf{0.848 $\pm$ 0.005}+ & \textbf{0.848 $\pm$ 0.005}+ & \textbf{0.850 $\pm$ 0.005}+& {0.878} & {0.879} & {0.879} & \underline{0.883} \\ \hline


    \multirow{4}{*}{Yeast$_2$} & -- & \textbf{0.961 $\pm$ 0.007}+  & \textbf{0.961 $\pm$ 0.007}+ & \textbf{0.960 $\pm$ 0.008}+ & \textbf{0.962 $\pm$ 0.007}+& {0.978} & {0.979} & {0.979} & {0.979} \\
   & 0.001 & \textbf{0.959 $\pm$ 0.008}+ & \textbf{0.958 $\pm$ 0.007}+ & \textbf{0.961 $\pm$ 0.006}+ & \textbf{0.961 $\pm$ 0.006}+ & {0.981} & {0.978} & {0.979} & {0.978} \\
     & 0.01     & {\textbf{0.955 $\pm$ 0.009}}+ & \textbf{0.959 $\pm$ 0.007}+ & \textbf{0.960 $\pm$ 0.009}+ & \textbf{0.961 $\pm$ 0.007}+ & {0.979} & {0.980} & {0.979} & {0.978} \\
      & 0.1     & \textbf{0.958 $\pm$ 0.009}+   & \textbf{0.960 $\pm$ 0.007}+ & \textbf{0.961 $\pm$ 0.007}+ & \textbf{\underline{0.962 $\pm$ 0.006}}+ & {0.978} & {0.978} & \underline{0.981} & {0.979} \\ \hline


    \multirow{4}{*}{Abal$_1$} & -- & 0.849 $\pm$ 0.081  & 0.862 $\pm$ 0.087  & 0.847 $\pm$ 0.089  &0.849 $\pm$ 0.085  & {0.964}  & 0.970  & 0.966  & 0.967 \\
   & 0.001 & \textbf{{0.892 $\pm$ 0.051}}+ &  \textbf{0.905 $\pm$ 0.036}+ & \textbf{0.907 $\pm$ 0.036}+ & \textbf{0.906 $\pm$ 0.034}+ & {0.970} & {0.968} & {0.969} & {0.971} \\ 
     & 0.01     & \textbf{0.908 $\pm$ 0.038}+ & \textbf{0.900 $\pm$ 0.056}+ & \textbf{\underline{0.919 $\pm$ 0.022}}+ & \textbf{0.919 $\pm$ 0.026}+ & {0.969} & \underline{0.973} & {0.970} & {0.972} \\
      & 0.1     & \textbf{0.910 $\pm$ 0.037}+ & \textbf{0.911 $\pm$ 0.046}+  & \textbf{0.912 $\pm$ 0.049}+ & \textbf{0.916 $\pm$ 0.031}+ & {0.970} & {0.972} & {0.969} & {0.970} \\  \hline

      \multirow{4}{*}{Abal$_2$} & -- &0.591 $\pm$ 0.102   &  0.623 $\pm$ 0.138  &0.634 $\pm$ 0.115   & 0.617  $\pm$ 0.137  & 0.862   & 0.878  & 0.881  &0.873  \\
       & 0.001 &   \textbf{0.729 $\pm$ 0.070}+ & \textbf{0.722 $\pm$ 0.063}+ & \textbf{{0.709 $\pm$ 0.080}}+ & \textbf{0.735 $\pm$ 0.074}+ & {0.877}  & {0.870} & {0.879}  & {0.885} \\
       & 0.01     & \textbf{0.721 $\pm$ 0.067}+ & \textbf{0.725 $\pm$ 0.075}+ & \textbf{0.721 $\pm$ 0.074}+ & \textbf{0.723 $\pm$ 0.066}+ & {0.881}  & {0.879}  & {0.884} & {0.880}\\ 
      & 0.1     &  \textbf{0.724 $\pm$ 0.076}+  & \textbf{0.739 $\pm$ 0.065}+ & \textbf{0.736 $\pm$ 0.063}+ & \textbf{\underline{0.756 $\pm$ 0.065}}+ & {0.888}  & {0.883} & {0.886} &  \underline{0.890}\\ \hline

      \multicolumn{2}{c}{Better (+) / Worse (-)  }  & \multirow{1}{*}{22 / 0  } & \multirow{1}{*}{22 / 0 } & \multirow{1}{*}{22 / 0 } & \multirow{1}{*}{22 / 0 }  \\

            \multicolumn{2}{c}{Same (=) / NSS}  & \multirow{1}{*}{ 0 / 2}  & \multirow{1}{*}{0 / 2} & \multirow{1}{*}{ 0 / 2 }  & \multirow{1}{*}{0 / 2 } \\

\hline

    \multicolumn{10}{c}{\textsf{NSGA-II PSDO}} \\ \hline
    \multirow{4}{*}{Ion} & -- & 0.794  $\pm$ 0.100  & \textbf{0.811 $\pm$ 0.084 }+ & 0.823  $\pm$ 0.091  & 0.795  $\pm$ 0.105  & {0.904 } & {0.932 } & {0.945 } & {0.939 }  \\
     & 0.001               & \textbf{{0.867 $\pm$ 0.035}}+ & \textbf{0.874 $\pm$ 0.029}+ & \textbf{0.880 $\pm$ 0.036}+ & \textbf{0.873 $\pm$ 0.045}+ & {0.959} & {0.952} & \underline{0.965} & {0.945} \\ 
     & 0.01     & \textbf{0.852 $\pm$ 0.050}+ & \textbf{0.867 $\pm$ 0.051}+ & \textbf{\underline{0.880 $\pm$ 0.031}}+ & \textbf{0.867 $\pm$ 0.050}+ & {0.947} & {0.950} & {0.944} & {0.949} \\
      & 0.1     & \textbf{0.853 $\pm$ 0.062}+ & \textbf{0.869 $\pm$ 0.048}+ &\textbf{0.875 $\pm$ 0.051}+ & \textbf{0.872 $\pm$ 0.049}+ & {0.941} & {0.951} & {0.956} & {0.938} \\ \hline 
    
   \multirow{4}{*}{Spect} & -- & \textbf{0.552 $\pm$ 0.020}+  & \textbf{0.546 $\pm$ 0.022}+  & \textbf{0.555 $\pm$ 0.022}+ & \textbf{0.554 $\pm$ 0.017}+ &  {0.648} & {0.665} & {0.638} & {0.640} \\
                       & 0.001 & \textbf{0.550 $\pm$ 0.026}+ & \textbf{0.562 $\pm$ 0.025}+ & \textbf{0.561 $\pm$ 0.025}+  & \textbf{0.592 $\pm$ 0.026}+ & {0.661} & {0.670} & {0.658} & \underline{0.706} \\ 
                       & 0.01  & \textbf{0.550 $\pm$ 0.025}+ & \textbf{0.563 $\pm$ 0.026}+ & \textbf{0.558 $\pm$ 0.025}+ & \textbf{0.583 $\pm$ 0.020}+ & {0.649} & {0.675} & {0.667} & {0.669} \\ 
                       & 0.1  & \textbf{0.551 $\pm$ 0.023}+ & \textbf{0.560 $\pm$ 0.024}+ & \textbf{0.557 $\pm$ 0.025}+ & \textbf{\underline{0.593 $\pm$ 0.020}}+ & {0.666} & {0.664} & {0.678} & {0.682}  \\  \hline

      \multirow{4}{*}{Yeast$_1$} & -- & \textbf{0.846 $\pm$ 0.006}+   &\textbf{0.846 $\pm$ 0.005}+    & \textbf{0.847 $\pm$ 0.005}+  &\textbf{0.848 $\pm$ 0.006}+   & {0.864}  & 0.868  & 0.871  &0.869  \\ 
      & 0.001 & \textbf{0.849 $\pm$ 0.006}+   & \textbf{0.848 $\pm$ 0.005}+  & \textbf{0.850 $\pm$ 0.007}+  & \textbf{0.850 $\pm$ 0.005}+  & 0.873  & 0.868  & 0.871  & 0.869  \\ 
     & 0.01     &  \textbf{0.850 $\pm$ 0.005}+  & \textbf{0.849 $\pm$ 0.007}+   & \textbf{0.850 $\pm$ 0.006}+  & \textbf{0.851 $\pm$ 0.006}+  & 0.870  & 0.874  & 0.872  & {0.872}  \\ 
      & 0.1     &  \textbf{0.850 $\pm$ 0.006}+  & \textbf{0.850 $\pm$ 0.005}+  & \textbf{0.850 $\pm$ 0.005}+ & \textbf{\underline{0.851 $\pm$ 0.006}}+  & \underline{0.876}  & 0.873  & 0.872  & 0.870  \\ \hline


      \multirow{4}{*}{Yeast$_2$} & -- & \textbf{0.957 $\pm$ 0.007}+  & \textbf{0.959 $\pm$ 0.007}+ & \textbf{0.957 $\pm$ 0.009}+ & \textbf{0.959 $\pm$ 0.007}+ & {0.973} & {0.978} & {0.976} & \underline{0.978} \\
   & 0.001 & \textbf{{0.960 $\pm$ 0.010}}+ & \textbf{0.962 $\pm$ 0.005}+ & \textbf{\underline{0.964 $\pm$ 0.005}}+ & \textbf{0.962 $\pm$ 0.008}+ & {0.976} & {0.976} & {0.978} & {0.977} \\
     & 0.01     & \textbf{0.962 $\pm$ 0.006}+ & \textbf{0.962 $\pm$ 0.006}+ & \textbf{0.962 $\pm$ 0.005}+ & \textbf{0.962 $\pm$ 0.006}+ & {0.977} & {0.975} & {0.974} & {0.975}  \\
  & 0.1     & \textbf{0.964 $\pm$ 0.006}+ & \textbf{0.960 $\pm$ 0.010}+ & \textbf{0.963 $\pm$ 0.005}+ & \textbf{{0.961 $\pm$ 0.007}}+ & {0.976} & {0.976} & {0.977} & {0.975} \\
  \hline
  


    \multirow{4}{*}{Abal$_1$} & -- & \textbf{\underline{0.890 $\pm$ 0.051}}+  & \textbf{0.881 $\pm$ 0.070}+  & \textbf{0.885 $\pm$ 0.046}+  &\textbf{0.884 $\pm$ 0.058}+  & {0.959}  & 0.966  & 0.961  & 0.952 \\
   & 0.001 & {{0.861 $\pm$ 0.079}} & {0.848 $\pm$ 0.073} & \textbf{{0.877 $\pm$ 0.078}}+ & {0.864 $\pm$ 0.075} & {0.962} & {0.957} & {0.962} & {0.959} \\ 
     & 0.01     & {0.864 $\pm$ 0.067} & {0.858 $\pm$ 0.076} & {{0.865 $\pm$ 0.070}} & 0.873 $\pm$ 0.066 & {0.967} & {0.959} & {0.962} & {0.962} \\
    & 0.1     & {0.858 $\pm$ 0.082} & \textbf{0.887 $\pm$ 0.061}+  & {0.864 $\pm$ 0.074} & {0.860 $\pm$ 0.075} & {0.963} & \underline{0.968} & {0.962} & {0.955} \\  \hline

      \multirow{4}{*}{Abal$_2$} & -- & \textbf{0.704 $\pm$ 0.083}+   &  \textbf{0.699 $\pm$ 0.072}+  & \textbf{0.706 $\pm$ 0.069}+   & \textbf{0.711  $\pm$ 0.076}+  & 0.826   & 0.859  & 0.874  &0.858  \\
       & 0.001 &   \textbf{0.725 $\pm$ 0.070}+ & \textbf{0.743 $\pm$ 0.079}+ & \textbf{{0.745 $\pm$ 0.060}}+ & \textbf{0.733 $\pm$ 0.075}+ & {0.859}  & {0.871} & {0.854}  & \underline{0.877} \\
       & 0.01     & \textbf{0.741 $\pm$ 0.086}+ & \textbf{0.735 $\pm$ 0.074}+ & \textbf{0.724 $\pm$ 0.070}+ & \textbf{0.728 $\pm$ 0.069}+ & {0.873}  & {0.867}  & {0.870} & {0.873}\\ 
      & 0.1     &  \textbf{0.743 $\pm$ 0.061}+  & \textbf{0.723 $\pm$ 0.073}+ & \textbf{0.719 $\pm$ 0.088}+ & \textbf{{0.722 $\pm$ 0.063}}+ & {0.877}  & {0.847} & {0.846} &  {0.851}\\ \hline

      \multicolumn{2}{c}{Better (+) / Worse (-) }  & \multirow{1}{*}{20 / 0} & \multirow{1}{*}{ 22/ 0} & \multirow{1}{*}{21 / 0} & \multirow{1}{*}{20 / 0} &   \\
      
                  \multicolumn{2}{c}{Same (=) / NSS}  & \multirow{1}{*}{ 0 / 4}  & \multirow{1}{*}{0 / 2} & \multirow{1}{*}{ 0 / 3 }  & \multirow{1}{*}{0 / 4} \\

\hline

 \hline     
 \multicolumn{10}{c}{\textsf{NSGA-II SSC}} \\ \hline

          \multirow{4}{*}{Ion}  & -- & 0.761 $\pm$ 0.108  &  0.749 $\pm$ 0.161  & 0.763 $\pm$ 0.152  & 0.744 $\pm$ 0.137  & 0.941  & 0.937  & 0.951  & 0.949 \\ 
      & 0.001 & 0.765 $\pm$ 0.134  & {0.753 $\pm$ 0.124 } & 0.699 $\pm$ 0.188  & 0.803 $\pm$ 0.103  & 0.954  & 0.935  & {0.928}  & 0.946   \\ 
       & 0.01     &  0.760 $\pm$ 0.125  & {0.751 $\pm$ 0.123}   & 0.710 $\pm$ 0.161  & 0.802 $\pm$ 0.104  & 0.947  & 0.929  & 0.928   & 0.947  \\
        & 0.1     &  {0.775 $\pm$ 0.095}  &  {0.738 $\pm$ 0.184}  & {0.746 $\pm$ 0.141}=  & {0.778 $\pm$ 0.099}  & {0.957}  & 0.951  & 0.945   & 0.936  \\ \hline

\multirow{4}{*}{Spect} & -- & 0.525 $\pm$ 0.025  & 0.532 $\pm$ 0.029  & 0.537 $\pm$ 0.020  & 0.535 $\pm$ 0.029  & 0.633  & 0.634  & 0.634  & 0.634  \\
      & 0.001 & 0.530 $\pm$ 0.029  &0.539 $\pm$ 0.030  &0.542 $\pm$ 0.023  & 0.540 $\pm$ 0.025  & 0.651  & 0.635  & 0.638  & 0.654  \\
     & 0.01     & 0.535 $\pm$ 0.029  &0.537 $\pm$ 0.027  & 0.541 $\pm$ 0.027  &0.540 $\pm$ 0.028   & 0.655  & 0.633  & {0.658}  & 0.651  \\
      & 0.1     & 0.532 $\pm$ 0.029  &0.531 $\pm$ 0.026  & 0.534 $\pm$ 0.027  & 0.533 $\pm$ 0.022  & {0.632}  & 0.641  & 0.635  & 0.635   \\ \hline 

      \multirow{4}{*}{Yeast$_1$} & -- & 0.819 $\pm$ 0.041   & 0.829 $\pm$ 0.023   & 0.835 $\pm$ 0.014  &0.834 $\pm$ 0.017   & {0.874}  & 0.875  & 0.878  &0.878  \\ 
      & 0.001 & 0.825 $\pm$ 0.031   &0.834 $\pm$ 0.029  &0.834 $\pm$ 0.019  &0.826 $\pm$ 0.039  & 0.877  & 0.877  & 0.877  & 0.877  \\ 
     & 0.01     &  0.827 $\pm$ 0.027  &0.835 $\pm$ 0.016   & 0.836 $\pm$ 0.019  & 0.830 $\pm$ 0.030  & 0.874  & 0.877  & 0.877  & {0.879}  \\ 
      & 0.1     &  0.831 $\pm$ 0.027  & 0.828 $\pm$ 0.034  &0.831 $\pm$ 0.028  &0.835 $\pm$ 0.014  & 0.879  & 0.876  & 0.876  & 0.875  \\ \hline



  \multirow{4}{*}{Yeast$_2$} & -- & 0.950 $\pm$ 0.013  &0.948 $\pm$ 0.010   & 0.945 $\pm$ 0.032  & 0.947 $\pm$ 0.009  &0.978   &0.977  &0.978  & 0.977  \\
 & 0.001 & 0.946 $\pm$ 0.013  &0.944 $\pm$ 0.028  &  0.947 $\pm$ 0.013  & 0.950 $\pm$ 0.011  & {0.976}  & 0.976  & 0.977  & {0.979}  \\
     & 0.01     & 0.947 $\pm$ 0.014  & 0.944 $\pm$ 0.024  & 0.946 $\pm$ 0.015  &0.949 $\pm$ 0.012  & 0.978  & 0.978  & 0.978  & 0.978   \\ 
      & 0.1     &0.948 $\pm$ 0.014   &0.948 $\pm$ 0.012  & 0.946 $\pm$ 0.009  & 0.947 $\pm$ 0.016  & 0.978  & 0.978  & 0.977  & 0.977  \\ \hline

        \multirow{4}{*}{Abal$_1$} & -- &  0.844 $\pm$ 0.084  & 0.839 $\pm$ 0.083  & 0.834 $\pm$ 0.070  &0.824 $\pm$ 0.099  &0.963   &0.967   &0.962  & 0.962  \\
     & 0.001 & 0.851 $\pm$ 0.062  & 0.812 $\pm$ 0.086  & 0.845 $\pm$ 0.077  &0.844 $\pm$ 0.079  & 0.964  & {0.961}  & 0.959  & 0.967  \\ 
     & 0.01     &  0.850 $\pm$ 0.076  & 0.833 $\pm$ 0.091  & 0.829 $\pm$ 0.096  & 0.836 $\pm$ 0.090  & {0.972}  & 0.957  & 0.959  & 0.963  \\
      & 0.1     &  0.869 $\pm$ 0.064  &0.838 $\pm$ 0.083  &0.844 $\pm$ 0.075  & 0.834 $\pm$ 0.084  & 0.963  & 0.965  & 0.965  & 0.962  \\ \hline


              \multirow{4}{*}{Abal$_2$} & -- & 0.521 $\pm$ 0.121   & 0.532 $\pm$ 0.103  & 0.529 $\pm$ 0.128  & \textbf{0.511 $\pm$ 0.118}-  & 0.810  & 0.802  & 0.841  & 0.801 \\
      & 0.001 & 0.561 $\pm$ 0.082  & {0.534 $\pm$ 0.102}  & 0.542 $\pm$ 0.104  & 0.502 $\pm$ 0.161  & 0.823  & {0.865}  & 0.829  & {0.820}  \\
     & 0.01     & 0.494 $\pm$ 0.147 & {0.536 $\pm$ 0.114}  & 0.533 $\pm$ 0.134  & 0.547 $\pm$ 0.123  & {0.844} & {0.826}  & 0.841  & 0.850  \\
      & 0.1     & \textbf{0.513 $\pm$ 0.132}- & 0.549 $\pm$ 0.120  & 0.514 $\pm$ 0.112 &0.532 $\pm$ 0.131  & {0.806} & 0.820  & {0.785} & 0.831  \\ \hline 

      \multicolumn{2}{c}{Better (+) / Worse (-)  }  & \multirow{1}{*}{0 / 1  } & \multirow{1}{*}{ 0 / 0 } & \multirow{1}{*}{0 / 0 } & \multirow{1}{*}{0 / 1 } & \multirow{1}{*}{ }   & \multirow{1}{*} { } & \multirow{1}{*}{ } & \multirow{1}{*}{ }  \\

            \multicolumn{2}{c}{Same (=) / NSS}  & \multirow{1}{*}{0 / 23}  & \multirow{1}{*}{ 0/ 24} & \multirow{1}{*}{1 / 23}  & \multirow{1}{*}{0 / 23}  & \multirow{1}{*}{ }    & \multirow{1}{*}{ }  &  \multirow{1}{*}{ } & \multirow{1}{*}{ }  \\

\hline

  \end{tabular}
  }
  \label{tab:nsgaii:semantics}
\end{table*}

\end{samepage}

\end{document}